\documentclass[10pt,twocolumn,letterpaper]{article}

\usepackage{cvpr}              % To produce the CAMERA-READY version
\definecolor{cvprblue}{rgb}{0.21,0.49,0.74}
\usepackage[pagebackref,breaklinks,colorlinks,allcolors=cvprblue]{hyperref}

\usepackage{tabularx}
\usepackage{graphicx}
\usepackage{booktabs}
\usepackage{multirow} 
\usepackage[dvipsnames]{xcolor}
\usepackage{array}
\newcolumntype{?}{!{\vrule width 0.7pt}}
\usepackage{makecell}

\def\paperID{351} % *** Enter the Paper ID here
\def\confName{CVPR}
\def\confYear{2025}

\title{PromptHMR: Promptable Human Mesh Recovery}

\author{\fontsize{12pt}{\baselineskip}\selectfont
        Yufu Wang\textsuperscript{1,4} \quad Yu Sun\textsuperscript{1} \quad Priyanka Patel\textsuperscript{1} \quad  Kostas Daniilidis\textsuperscript{4,5} \\ 
        \fontsize{12pt}{\baselineskip}\selectfont
        Michael J. Black\textsuperscript{1,2} \quad  Muhammed Kocabas\textsuperscript{1,2,3}\\
        \fontsize{11pt}{\baselineskip}\selectfont
        $^1$ Meshcapade ~
        $^2$ MPI for Intelligent Systems ~
        $^3$ ETH Zürich ~
        $^4$ University of Pennsylvania ~
        $^5$ Archimedes ~\\
         \url{https://yufu-wang.github.io/phmr-page}
    }

\begin{document}
% \maketitle
\twocolumn[\maketitle\vspace{-2mm}\begin{center}
    \vspace{-2em}
    \includegraphics[width=0.915\linewidth]{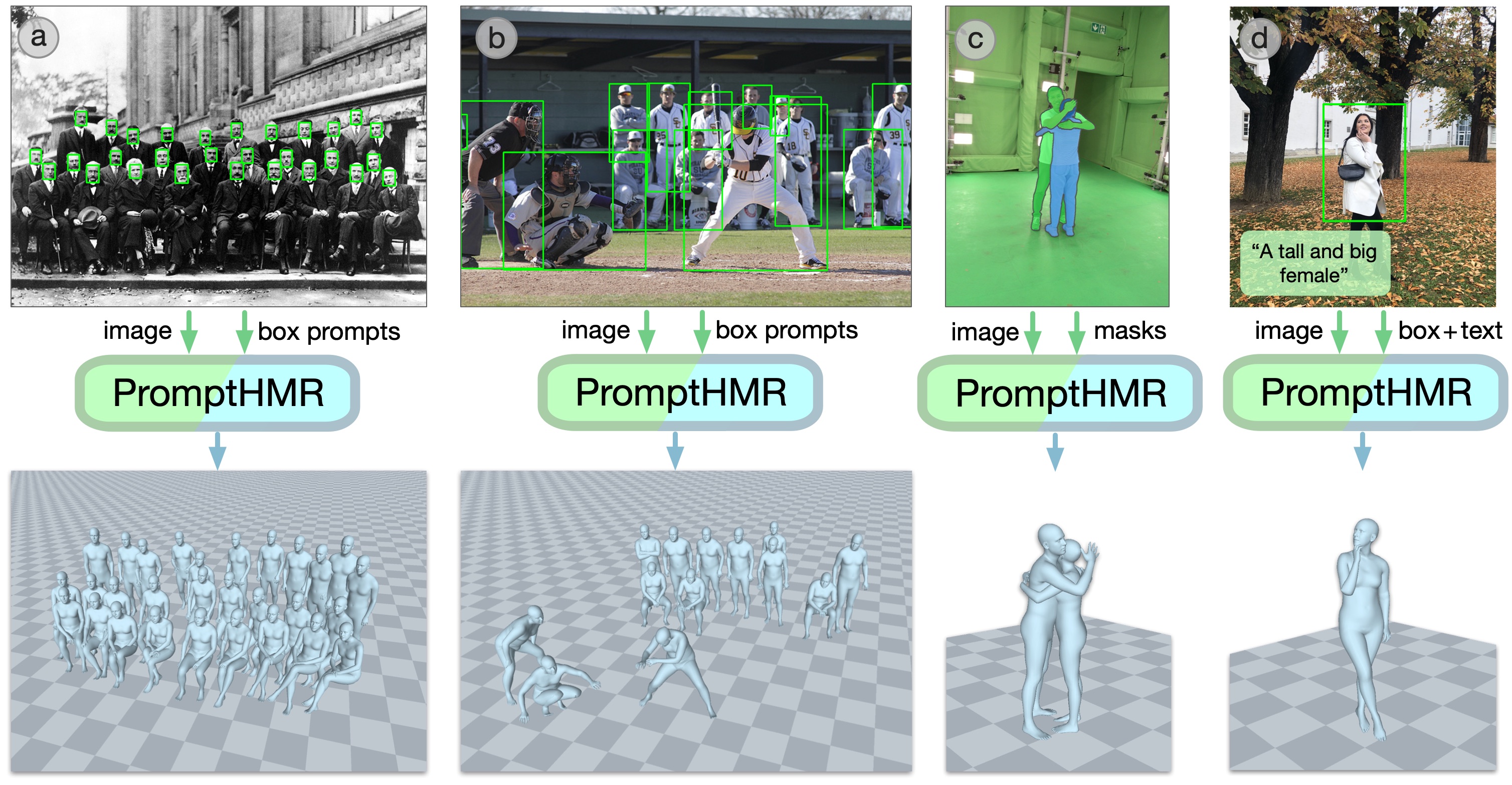}
    \vspace{-2mm}
    \captionof{figure}{\textbf{PromptHMR} is a promptable human pose and shape (HPS) estimation method that processes images with spatial or semantic prompts. It takes ``side information" readily available from vision-language models or user input to improve the accuracy and robustness of 3D HPS.
    PromptHMR recovers human pose and shape from spatial prompts such as (a) face bounding boxes, (b) partial or complete person detection boxes, or (c) segmentation masks. It refines its predictions using semantic prompts such as (c) person-person interaction labels for close contact scenarios, or (d) natural language descriptions of body shape to improve body shape predictions. Both image and video versions of PromptHMR achieve state-of-the-art accuracy. }
    \label{fig:teaser}
\end{center}
\vspace{-2mm}\vspace{2mm}\bigbreak]
\begin{abstract}
Human pose and shape (HPS) estimation presents challenges in diverse scenarios such as crowded scenes, person-person interactions, and single-view reconstruction. Existing approaches lack mechanisms to incorporate auxiliary ``side information" that could enhance reconstruction accuracy in such challenging scenarios. Furthermore, the most accurate methods rely on cropped person detections and cannot exploit scene context while methods that process the whole image often fail to detect people and are less accurate than methods that use crops. While recent language-based methods explore HPS reasoning through large language or vision-language models, their metric accuracy is well below the state of the art. In contrast, we present PromptHMR, a transformer-based promptable method that reformulates HPS estimation through spatial and semantic prompts. Our method processes full images to maintain scene context and accepts multiple input modalities: spatial prompts like bounding boxes and masks, and semantic prompts like language descriptions or interaction labels. PromptHMR demonstrates robust performance across challenging scenarios: estimating people from bounding boxes as small as faces in crowded scenes, improving body shape estimation through language descriptions, modeling person-person interactions, and producing temporally coherent motions in videos. Experiments on benchmarks show that PromptHMR achieves state-of-the-art performance while offering flexible prompt-based control over the HPS estimation process.
\end{abstract}
\vspace{-10mm}    
\section{Introduction}
\label{sec:intro}
The estimation of 3D human pose and shape (HPS) is classically viewed as regressing the parameters of shape and pose from pixels.
In particular, most methods take a tightly cropped image of a person and output the pose and shape in camera coordinates.
While the accuracy of such methods has increased rapidly, they do not address the whole problem.
In particular, an HPS method should be able to take an image or video containing complex human-human and human-scene interactions, return the parameters of every person in the scene, and place these people in a consistent global coordinate frame.

Our key observation is that the classical ``pixels to parameters" formulation of the problem is too narrow.
Today, we have large vision-language foundation models (VLMs) that understand a great deal about images and what people are doing in them.
What these models lack, however, is an understanding of 3D human pose and shape.
Recent work \cite{feng2024chatpose,delmas2024poseembroider} has tried to bring together VLMs and 3D HPS but with 3D accuracy well below the best classical methods.

Consequently, we need to think about the problem in a different way and ask whether we can exploit readily available side information (e.g.~provided by a VLM) to improve 3D HPS regression robustness, usefulness, and accuracy.
To that end, we develop a novel ``promptable'' HPS architecture called \textbf{PromptHMR}.
Consider the sample images shown in Fig.~\ref{fig:teaser}. 
In crowded scenes, existing person detection methods struggle, while face detection methods remain reliable. 
When people closely interact, their body parts overlap and occlude each other, introducing ambiguity in pose estimation. 
Moreover, 3D body shape estimation from monocular views is challenging due to perspective ambiguity. 
In all these cases, we can extract cues, or prompts, that provide ``side information'' that can help an HPS method better analyze the scene.
PromptHMR formalizes this intuition by combining image evidence with different types of spatial and semantic information that can come from either humans or AI systems such as VLMs.

Specifically, our approach combines three key components: 
(1) a vision transformer that extracts features from high-resolution full images to preserve scene context, 
(2) a multi-modal prompt encoder that processes spatial and semantic inputs, and 
(3) a transformer decoder that attends to both prompt and image tokens to generate SMPL-X \cite{smplx} body parameters. 
This design addresses the limitations of cropped-image HPS methods by processing full images using side information in the form of prompts. 
It addresses the challenges that full-image HPS methods have in detecting all people in a scene by accepting readily available bounding boxes. 
Last, our method incorporates auxiliary semantic information through text descriptions or interaction labels. 

By combining spatial and semantic prompting, our method offers a powerful and versatile approach to 3D HPS estimation from the whole image. At test time, we show that this promptable structure (1) can take various bounding boxes or segmentation masks to recover full body HPS in a robust way, (2) improve its body shape predictions by using textual descriptions as input, (3) is capable of modeling person-person close interaction directly in the regression process, and (4) uses full image context to reconstruct people coherently in the camera space and the world space. 
Our model can handle video by incorporating temporal transformer layers at the SMPL-X decoding phase, yielding temporally stable and smooth motions. 
Last, following TRAM~\cite{wang2024tram}, we combine the temporal version of our model with metric SLAM to estimate human motion in world coordinates. 

We make several key design choices that make Prompt\-HMR successful. 
To achieve robustness to different spatial inputs, we train our model by simulating noisy full-body and face-region bounding boxes. 
For improved body shape estimation, we leverage SHAPY~\cite{shapy} to generate automatic body shape descriptions for training samples and process them with a pretrained text encoder~\cite{radford2021learning}. 
To enhance person-person interaction reconstruction, we use segmentation masks as more precise spatial prompts and develop person-person attention layers that operate between prompted people, producing coherent reconstructions of close interactions. 
Through random masking of different input types during training, our model learns to work with any combination of prompts at test time. 

Quantitative experiments on the EMDB~\cite{emdb}, 3DPW~\cite{3dpw}, RICH~\cite{rich}, Hi4D~\cite{hi4d}, CHI3D~\cite{chi3d} and HBW~\cite{shapy} benchmark datasets demonstrate that our method outperforms state-of-the-art (SOTA) approaches and strong baselines. 
We also provide many qualitative examples of in-the-wild images and videos that illustrate the robustness and generalization of PromptHMR.

By moving away from the pure pixels-to-parameters approach, PromptHMR not only achieves a new SOTA, it shows a new way of improving both accuracy and robustness by leveraging side information that is easily available.
One can think of this as a collaboration between VLMs, which know a lot about people in images but not in 3D, and a metric regressor that knows a lot about 3D humans but not about the semantics of what they do.
We show that this combination has significant upside potential to increase both generality and accuracy.
Our code and model are available for research purposes.
\section{Related Work}
\label{sec:related}

\textbf{Human pose and shape estimation from images}.
Existing methods for human pose and shape (HPS) estimation can be broadly categorized into two main approaches. The first \cite{hmr,spin,refit,pare,hmr2,pymaf,cliff,I2l,pose2mesh, SPEC:ICCV:2021, meshgraphform} uses a tightly cropped image of an individual as input, and estimates pose and shape in camera coordinates. While effective for isolated individuals, this approach discards scene context that is essential to resolve human pose in cases of occlusion, severe overlap and close interaction in multi-person scenes~\cite{hi4d, chi3d}.

The second category \cite{multipeople,romp,bev,trace,multihmr,sun2024aios} build upon object detection frameworks~\cite{centernet, detr} to jointly detect humans and estimate their pose and shape parameters. Having access to the entire image, they can better perceive occluded individuals and infer depth relationships, but they often suffer from detection failures due to the difficulty in simultaneously performing detection and reconstruction. Our ``promptable" architecture leverages detection box prompts to resolve such conflicts while having access to the entire scene context.

\noindent\textbf{Human pose and shape estimation from video.}
Methods for human motion estimation from video can also be divided into two main categories. 
The first \cite{vibe,dsd,hmmr,Luo_2020_ACCV,tcmr} focuses on estimating smooth human motion in camera space. These methods build upon single-person HPS estimation approaches \cite{hmr,spin} by adding temporal layers during the SMPL decoding phase to introduce temporal coherence.

More recent methods estimate human motion in world coordinates from videos captured with dynamic cameras. These methods follow a two-stage approach, first estimating camera motion using SLAM techniques \cite{droid,orb,orb2,bodyslam,bodyslam++,teed2024dpvo}, and then leveraging human motion priors to optimize the human world motion \cite{pace,slahmr,glamr}. Others \cite{wham, shen2024gvhmr} learn temporal models to directly regress human world motion from image and camera features. Still others \cite{wang2024tram,Zhao2024globalcamhuman} use monocular metric depth estimation to solve for the scale of camera motion and transform human motion from camera space to world coordinates.

In our approach, we extend PromptHMR to video by taking the SMPL-X output tokens and utilizing a temporal transformer module to estimate temporally stable and smooth human motion and translation in camera space. We follow TRAM \cite{wang2024tram} to transform human motion to world coordinates due to its simplicity and effectiveness.

\noindent\textbf{Semantic reasoning about 3D humans in images.} 
Recent methods explore combining different types of semantic information, such as language descriptions and knowledge of person-person interactions, to improve reasoning about 3D humans from images and videos.
For example, ChatPose~\cite{feng2024chatpose} follows the common approach of visual language models (VLMs) \cite{liu2023llava} by fine-tuning a large language model (LLM) with a combination of images and tokens to estimate SMPL parameters. 
In a similar direction, PoseEmbroider \cite{delmas2024poseembroider} is a multi-modal framework that aligns image, 3D pose, and text representations in a shared latent space.
While ChatPose focuses on combining high-level scene reasoning with 3D HPS, PoseEmbroider exploits detailed language descriptions of human pose.
While promising, neither method achieves SOTA accuracy on the HPS task.
Note that many other methods relate language to human pose or motion, without considering images \cite{posefix,posegpt,motionfix,mdm,actor}, but these are outside our scope.

Additionally, several methods \cite{R-ARBOL_2024_ECCV_BodyShapeGpt,bodytalk,shapy} focus on modeling the relationship between SMPL body shape and natural language descriptions. These methods show that language descriptions and images can provide complementary information to solve this task.
Other approaches, such as BUDDI \cite{buddi} and ProsePose \cite{ProsePose}, address the challenge of estimating person-person interactions. BUDDI is an optimization-based approach that leverages diffusion model as a prior over interacting people, while
ProsePose queries a VLM to estimate contact points on the human body surface and uses these contact points to guide an optimization process for improving human interaction.

Overall, methods like ChatPose~\cite{feng2024chatpose} and PoseEmbroider~\cite{delmas2024poseembroider} are promising steps toward jointly learning the relationship between vision, language, and 3D humans, but their understanding of 3D humans remains limited, as indicated by their relatively low 3D pose accuracy. Meanwhile, SHAPY~\cite{shapy}, BodyShapeGPT~\cite{R-ARBOL_2024_ECCV_BodyShapeGpt}, and BodyTalk~\cite{bodytalk} focus solely on exploring the relationship between SMPL body shape and natural language. BUDDI and ProsePose are post-processing approaches for interaction that do not directly reason using image information.

Our approach addresses the limitations of these methods by training a single model capable of flexible prompting that achieves state-of-the-art (SOTA) performance, not only on standard HPS benchmarks but also on benchmarks tailored to body shape and person-person interaction.
\begin{figure*}[h!]
    \centering
    \includegraphics[width=0.99\textwidth]{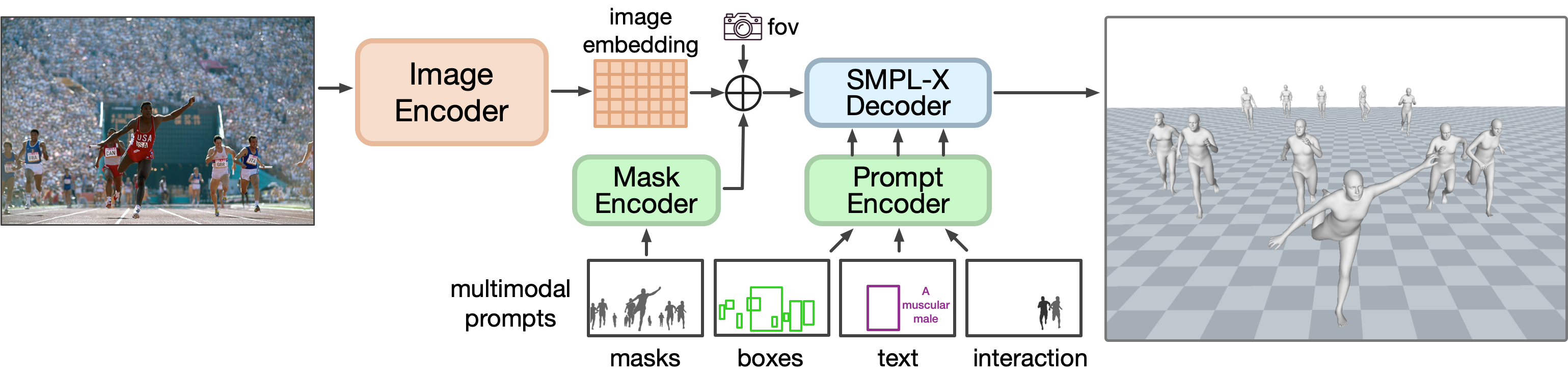}
    \vspace{-2mm}
    \caption{\textbf{Method overview.} PromptHMR estimates SMPL-X parameters for each person in an image based on various types of prompts, such as boxes, language descriptions, and person-person interaction cues. Given an image and prompts, we utilize \textcolor{Orange}{a vision transformer} to generate image embeddings and \textcolor{LimeGreen}{mask and prompt encoders} to map different types of prompts to tokens. Optionally, camera intrinsics can be embedded along with the image embeddings. The image embeddings and prompt tokens are then fed to the \textcolor{SkyBlue}{SMPL-X decoder}. The SMPL-X decoder is a transformer-based module that attends to both the image and prompt tokens to estimate SMPL-X parameters. Note that the language and interaction prompts are optional, but providing them enhances the accuracy of the estimated SMPL-X parameters.}
    \label{fig:method}
    \vspace{-2mm}
\end{figure*}

\section{Method}
\label{sec:method}

Given an image $I$ containing $N$ people and a set of prompts, our main goal is to recover the pose, shape, and locations of the people in the camera space to form a coherent human-centric 3D scene. Figure~\ref{fig:method} shows an overview.

\subsection{Promptable mesh regression}
\label{sec:architecture}
We adopt SMPL-X~\cite{smplx} to represent each person $i$ in the 3D space, including the orientation $\phi_i \in \mathbb{R}^{3}$, local body pose $\theta_i \in \mathbb{R}^{22\times3}$, shape $\beta_i \in \mathbb{R}^{10}$, and translation $\tau_i \in \mathbb{R}^{3}$ in the camera space. 
We do not include face and hand parameters in this work. 
Each human $H_i$ is mapped to a 3D mesh with the differentiable SMPL-X layer. 
\begin{equation}
    H_i = \{ \phi_i, \theta_i, \beta_i, \tau_i \}.
    \label{eq:smplx}
\end{equation}

Each person can be prompted with spatial and semantic prompts. Spatial prompts include a bounding box $b_i \in \mathbb{R}^{2\times2}$ (the two corners) and a segmentation mask $m_i \in \mathbb{R}^{h\times w}$. Semantic prompts consist of text and two-person interaction labels. The text prompt is the CLIP embedding $t_i$ of a sentence describing the body shape. The interaction prompt is a binary variable $k_i$ indicating whether two people are in close contact. While semantic prompts are optional, each human needs at least one spatial prompt to be reconstructed. Overall, the input prompts are represented as $P_i$:
\begin{equation}
\begin{split}
    P_i \subseteq \{ b_i, m_i, t_i, k_i \}   \\
    b_i \in P_i \text{ or } m_i \in P_i
    \label{eq:prompts}
\end{split}
\end{equation}
Promptable human mesh recovery (PromptHMR) is defined as a learnable function that maps an image and a set of prompts to a set of 3D humans
\begin{equation}
    f: (I, \{P_i\}_{i=1}^N)\rightarrow \{ H_i \}_{i=1}^N .
    \label{eq:prompthmr}
\end{equation}
This task definition integrates all available contexts to locate and reconstruct prompted humans in the image. 

\subsection{Model}
\label{sec:model}
\textbf{Image encoder.} The image is first encoded as tokens by a vision transformer (ViT) encoder from DINOv2~\cite{dosovitskiy2020vit, oquab2023dinov2}:
\begin{equation}
    F = \mathrm{Encoder}(I),
    \label{eq:encoder}
\end{equation}
To ensure sufficient resolution for modeling humans at both near and far distances, we use $896\times896$ images. The encoder is run once per frame regardless of the number of people prompted. When camera intrinsics are provided, we add positional encoding of the camera rays to the image tokens to make them camera-aware~\cite{multihmr, Facil_2019}.

\noindent\textbf{Mask encoder.} When available, masks are first processed by an encoder consisting of strided convolutional layers that downsample the masks. The output mask features are added to the image tokens. If no mask is provided, a learned ``no mask" token is added instead. 
\begin{equation}
    F_i = \mathrm{Encoder}_{\text{m}}(m_i) + F .
    \label{eq:encoder}
\end{equation}

\noindent\textbf{Prompt encoder.} The prompt encoder consists of a set of transformations that map different types of prompts to token vectors of the same dimension. When a prompt is not available, it is replaced with a learned null token.

For bounding boxes, we encode $b_i$ using positional encoding summed with learned embeddings to form the box prompt tokens $T_{bi} = \mathrm{PE}(b_i)$, with $T_{bi} \in \mathbb{R}^{2\times d}$.  We design different box transformations during training to allow the model to use different boxes as a human identifier. In the training phase, each instance is prompted with either a whole-body bounding box, a face bounding box, or a truncated box covering part of the body. Gaussian noise is added to both corners. At inference time, the model accepts boxes without needing to know the box types.   

Language is a natural way to supply semantic information, and in this paper, we use language to supplement spatial prompts with information on body shape. A sentence such as ``a muscular and tall male" is encoded with the CLIP text encoder $T_{ti} = \mathrm{CLIP}(t_i)$, with $T_{ti} \in \mathbb{R}^{d}$. To generate paired (image, text) data, we run SHAPY's~\cite{shapy} shape-to-attribute method on the ground truth shape parameters to obtain shape attribute scores and randomly pick a subset of top attributes to form a sentence. 

The interaction prompt $k_i$ passes through the prompt encoder without modification and directly switches on-off the cross-person attention that is described in Sec.~\ref{interaction}.

\noindent\textbf{SMPL-X decoder.} The SMPL-X decoder appends two query tokens $T_{\text{smpl}}, T_\text{depth}$ with the prompt tokens $T_{bi}, T_{ti}$ to form the person-specific prompt $T_i \in \mathbb{R}^{5\times d}$.

Finally, we use a standard transformer decoder and two MLP heads to produce the final output
\begin{equation}
\begin{split}
    T^{\prime}_{smpl}, T^{\prime}_{depth} & = \mathrm{Decoder}(F_i, T_i)  \\
    \phi_i, \theta_i, \beta_i & = \mathrm{Head}_{smpl}(T^{\prime}_{smpl}) \\
    \tau_i & = \mathrm{Head}_{depth}(T^{\prime}_{depth}).
    \label{eq:decoder}
\end{split}
\end{equation}

The transformer consists of three attention blocks. Each block applies self-attention on the tokens, cross-person attention (described in Sec.~\ref{interaction}), and then two-way cross-attention between the tokens and the image embeddings~\cite{sam}. The self-attention and cross-attention with the image are applied to each prompted person independently. We use separate tokens $T_{\text{smpl}}$ and $T_\text{depth}$ to make the location representation invariant to the 3D human pose and shape representation. 

Regressing the location of the human in the camera space is much more challenging than most prior work that models humans in a cropped image space. Therefore, we do not regress $\tau$ directly. We regress focal length normalized 2D translation $p_{xy}\in \mathbb{R}^2$ and inverse depth $p_z \in \mathbb{R}$, and then transform them to $\tau$ as follows
\begin{equation} 
    t_{xy} = \frac{p_{xy}}{p_z} \quad
    t_z = \frac{1}{p_z} \times \frac{f}{f_c} \quad
    \tau = [t_{xy}, tz],
    \label{eq:transl}
\end{equation}
where $f$ is the ground truth or estimated focal length of the image, and $f_c$ is the canonical focal length. Predicting the normalized inverse depth follows the recent monocular depth literature~\cite{ranftl2022midas} and is also intuitive since the inverse depth is linearly related to the size of the human in the image. Predicting $p_{xy}$ is equivalent to predicting the 2D location of the human in a normalized image plane.

\subsection{Two-person interaction}
\label{interaction}
We introduce promptable layers in the decoder to model two-person interaction. We describe the case where there are two people in the image, but the implementation can extend to model an interacting pair in a larger group. 

\begin{figure}[t!]
    \centering
    \includegraphics[width=0.47\textwidth]{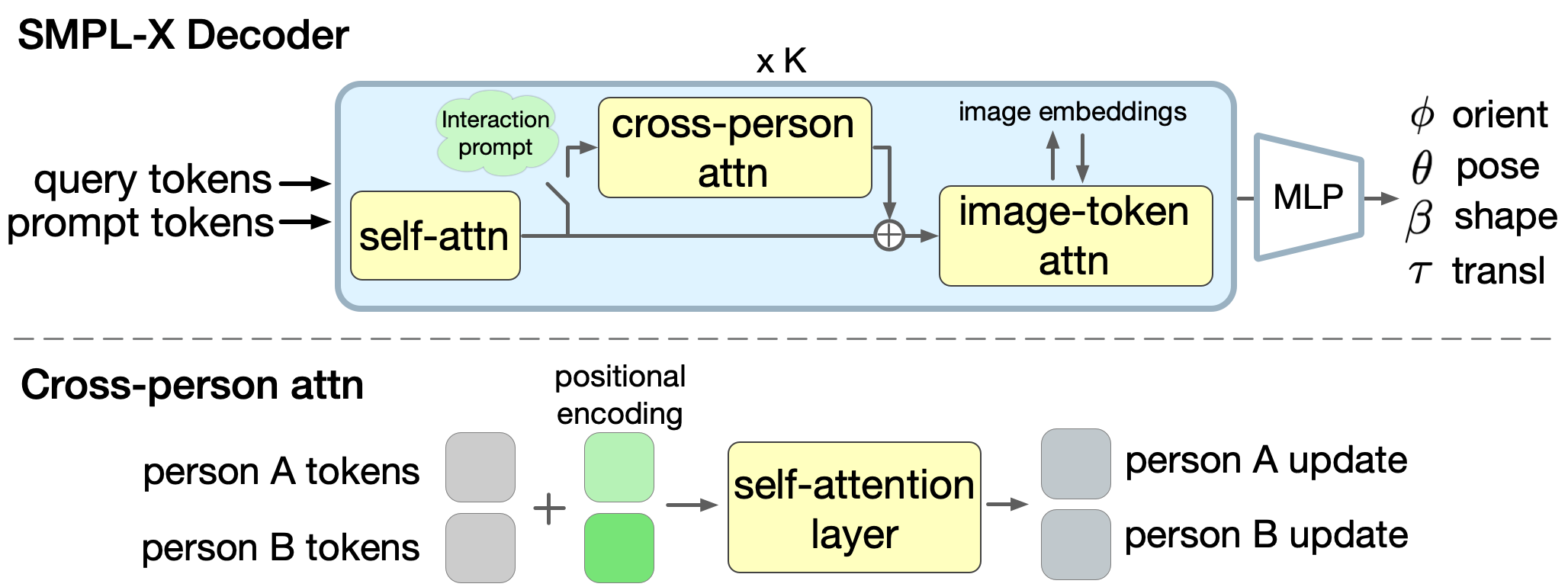}
    \vspace{-1mm}
    \caption{\textbf{SMPL-X decoder.} The top row shows one attention block in the decoder. The cross-person interaction module can be turned on/off. The bottom row shows the cross-person attention.}
    \label{fig:smplx_decoder}
    \vspace{-3mm}
\end{figure}

The promptability is modeled as a flow control with a residual connection (Fig.~\ref{fig:smplx_decoder}). Specifically, if two humans are interacting (as indicated by $k_i$), their query tokens pass through an additional self-attention layer; otherwise, non-interacting humans skip this. 

Applying attention to every person often creates unnecessary dependency in crowded scenes, and there is limited training data for large-group scenarios. However, there is high-quality data featuring two-person social interactions. By making the interaction layers promptable, we mitigate data diversity issues and increase flexibility, regardless of the number of people in the scene.

Our proposed interaction layer uses a standard self-attention mechanism. First, we add positional encodings to the query tokens to distinguish the two individuals. The encoded tokens then go through a self-attention layer, whose output is combined with the original tokens via a residual connection. Our experiments demonstrate that including these interaction layers significantly improves inter-person pose accuracy in two-person interaction benchmarks.

\subsection{PromptHMR video version}
In addition to the single-image variant of PromptHMR, we train an extended version that processes videos to estimate human motion in world coordinates. To achieve this, we introduce a simple and efficient temporal transformer module. Given a monocular video sequence $\{I^t\}_{t=0}^T$, we first run PromptHMR to obtain per-subject SMPL-X decoder output tokens $T^{\prime}_\text{smpl}$ and $T^{\prime}_\text{depth}$, assuming that the subject identities are provided with the prompts. These tokens, along with the positional encoding of time $t$, are fed to a decoder-only temporal transformer module with twelve attention blocks. The output tokens are converted to SMPL-X parameters ${ \phi_t, \theta_t, \beta_t }$, translation $\tau_t$, and joint contact probabilities $c_t$. The contact probabilities indicate whether a given joint is in contact with the ground plane similar to ~\cite{humor,wham,shen2024gvhmr}.

To obtain results in world coordinates, we adopt the approach from TRAM~\cite{wang2024tram}. Specifically, we use DROID-SLAM~\cite{droid} and a monocular metric depth estimation model, ZoeDepth~\cite{zoedepth}, to estimate camera motion in metric world coordinates. The translation parameters $\tau_t$ are then transformed to world coordinates using the estimated camera motion. To refine the human trajectory and mitigate foot-skating artifacts, we leverage the estimated contact probabilities and run a fast postprocessing that optimizes the contact joints to have zero velocity.

\subsection{Losses}
\label{sec:losses}
PromptHMR is trained with a combination of 2D and 3D losses, following traditional HMR methods~\cite{hmr,spin}:
\begin{align*}
\mathcal{L} = \lambda_{1}\mathcal{L}_{2D} + \lambda_{2}\mathcal{L}_{3D} + \lambda_{3}\mathcal{L}_{\mathrm{SMPL}} + \lambda_{4}\mathcal{L}_{V} + \lambda_{5}\mathcal{L}_{\mathit{t}}
\label{eq:loss}
\end{align*}
with each term calculated as 
\begin{align*}
\mathcal{L}_{2D} &= ||\mathcal{\hat{J}}_{2D} - \Pi(\mathcal{J}_{3D})||^2_F \\
\mathcal{L}_{3D} &= ||\mathcal{\hat{J}}_{3D} - \mathcal{J}_{3D}||^2_F \\
\mathcal{L}_{\mathrm{SMPL}} &= ||\hat{\Theta} - \Theta||^2_2 \\
\mathcal{L}_{V} &= ||\hat{V} - V||^2_F \\
\mathcal{L}_{\mathit{t}} &= ||\hat{p}_{xy} - p_{xy}||^2_F + ||\hat{p}_z - p_z||^2_F
\end{align*}
where $\mathcal{J}_{3D}$ and $V$ are the 3D joints and vertices of the SMPL-X model, with the hat operator denoting the ground truth. $\Pi$ is the camera reprojection operator. Additionally, on datasets with ground truth translation labels, we supervise the normalized translation $p_{xy}$ and inverse depth $p_z$.

\section{Experiments}
\label{sec:exp}
\textbf{Datasets.} We train PromptHMR with standard datasets: BEDLAM~\cite{bedlam}, AGORA~\cite{agora}, 3DPW~\cite{3dpw}, COCO~\cite{coco}, and MPII~\cite{mpii3d}. Following 4DHumans, we add AIC~\cite{aic} and InstaVariety~\cite{hmmr} as in-the-wild data, with pseudo-ground truth from CamSMPLify~\cite{camerahmr}. Additionally, we add CHI3D~\cite{chi3d} and HI4D~\cite{hi4d} to enable learning two-person interaction following the train-test splits from BUDDI~\cite{buddi}. Including CHI3D and HI4D does not improve performance on other benchmarks. 

\noindent\textbf{Implementation.} We train PromptHMR with AdamW with a batch size of 96 images of resolution 896$\times$896. We use a learning rate of $1e^{-5}$ for the image encoder and $3e^{-5}$ for the prompt encoder and the SMPL-X decoder, with a weight decay of $5e^{-5}$. The training converges within 350K steps.

\noindent\textbf{Evaluation.} We evaluate camera space reconstruction accuracy on 3DPW~\cite{3dpw}, EMDB~\cite{emdb} and RICH~\cite{rich}, using MPJPE, Procrustes-aligned MPJPE (PA-MPJPE) and Per Vertex Error (PVE)~\cite{hmr}. We evaluate inter-person accuracy on HI4D and CHI3D by Pair-PA-MPJPE, which aligns the two people as a whole with the ground truth~\cite{buddi}.

To evaluate world-grounded motion on EMDB with PromptHMR video (PromptHMR-vid), we compute World-aligned MPJPE (WA-MPJPE$_{100}$), World MPJPE (W-MPJPE$_{100}$) and Root Translation Error (RTE in $\%$)~\cite{slahmr, wham}.

\subsection{Reconstruction accuracy}
For camera space reconstruction, as shown in Table~\ref{tab:main_image}, PromptHMR and PromptHMR-Vid demonstrate state-of-the-art performance, matching crop-based methods while achieving better results than other full-image methods. PromptHMR and CameraHMR use the same training data and have similar performance, which validates that this prompt-based approach can achieve metrically accurate results. 
For representative results, see Fig.~\ref{fig:results}, where PromptHMR recovers coherent 3D scenes of people.

\begin{figure*}[h!]
    \centering
    \includegraphics[width=0.99\textwidth]{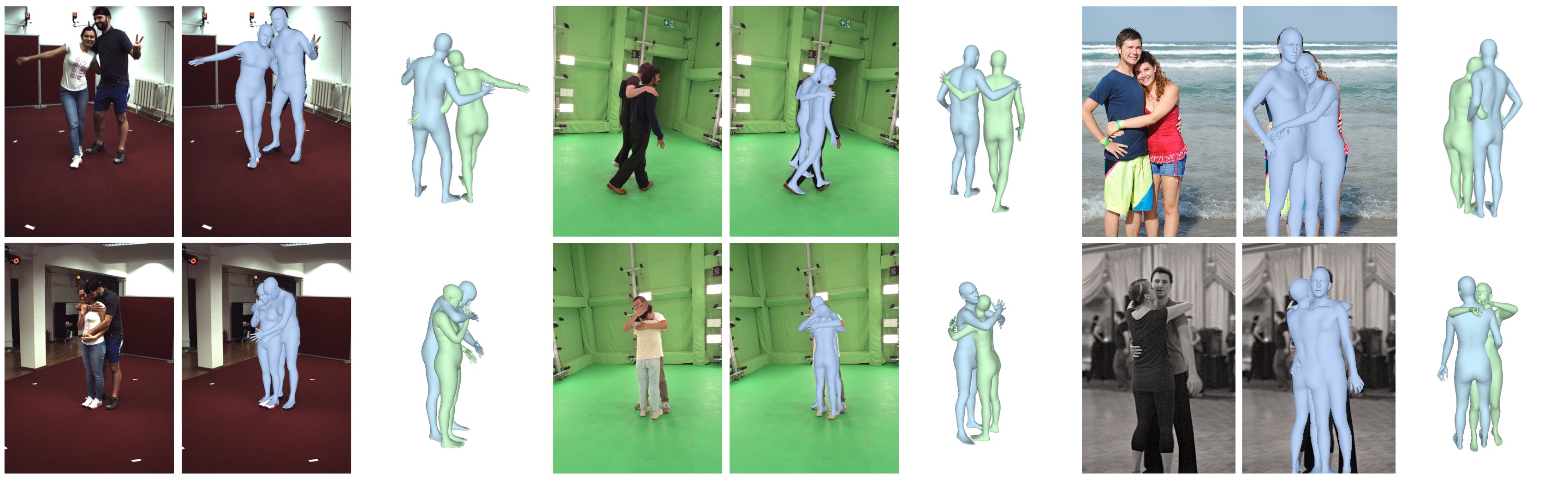}
    \vspace{-2mm}
    \caption{\textbf{Qualitative results.} PromptHMR recovers coherent two-person close interaction. 
    Despite suffering from some interpenetration, the relative positions of the interacting people are accurately recovered. More examples are provided in the Supplementary.}
    \vspace{-2mm}
    \label{fig:interaction}
\end{figure*}

For interaction reconstruction, PromptHMR achieves good accuracy as indicated in Table~\ref{tab:main_interaction}. Compared to BUDDI which is also trained on CHI3D and HI4D, our method achieves better overall accuracy on per-person and inter-person metrics. We show qualitative results in Fig.~\ref{fig:interaction}. As a monocular regression method, PromptHMR still cannot avoid interpenetration between closely interacting people. 

PromptHMR-Vid achieves SOTA performance among methods that estimate human motion in world coordinates, as shown in Table~\ref{tab:world_coord_eval}. Unlike TRAM, we estimate the joint contact probabilities similar to \cite{wham,shen2024gvhmr}. Therefore, we achieve lower foot skating than TRAM, even though we use the same metric SLAM method to transform motion in camera space to world coordinates. Please refer to our supplementary material (SupMat) for qualitative results of PromptHMR-Vid.

\begin{figure}[ht!]
    \centering
    \includegraphics[width=0.47\textwidth]{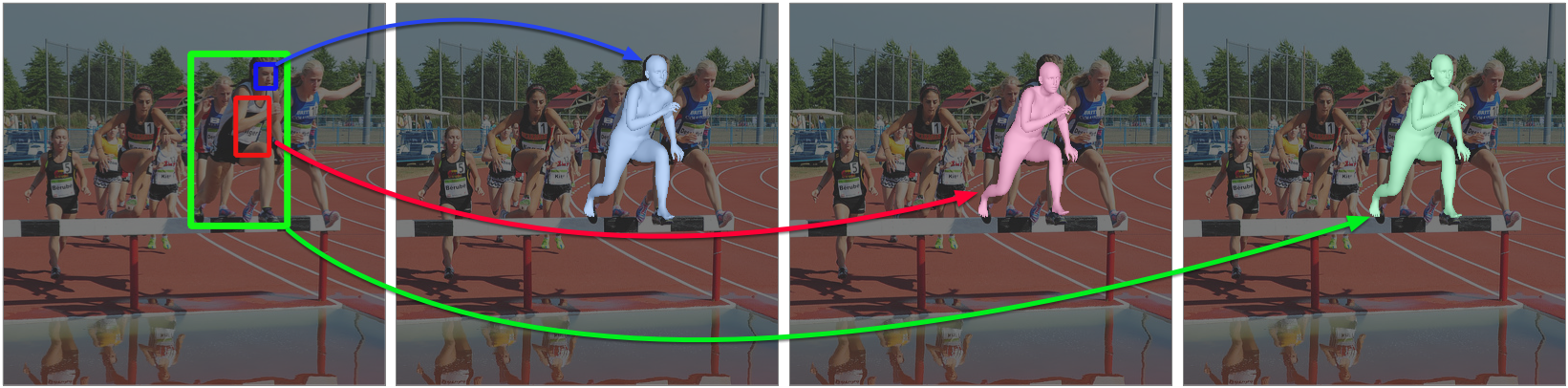}
    \vspace{-2mm}
    \caption{\textbf{Effect of box prompts.} Our method remains stable with different boxes, including noisy truncated boxes.}
    \label{fig:boxes}
    \vspace{-2mm}
\end{figure}

\begin{figure}[ht!]
    \centering
    \includegraphics[width=0.47\textwidth]{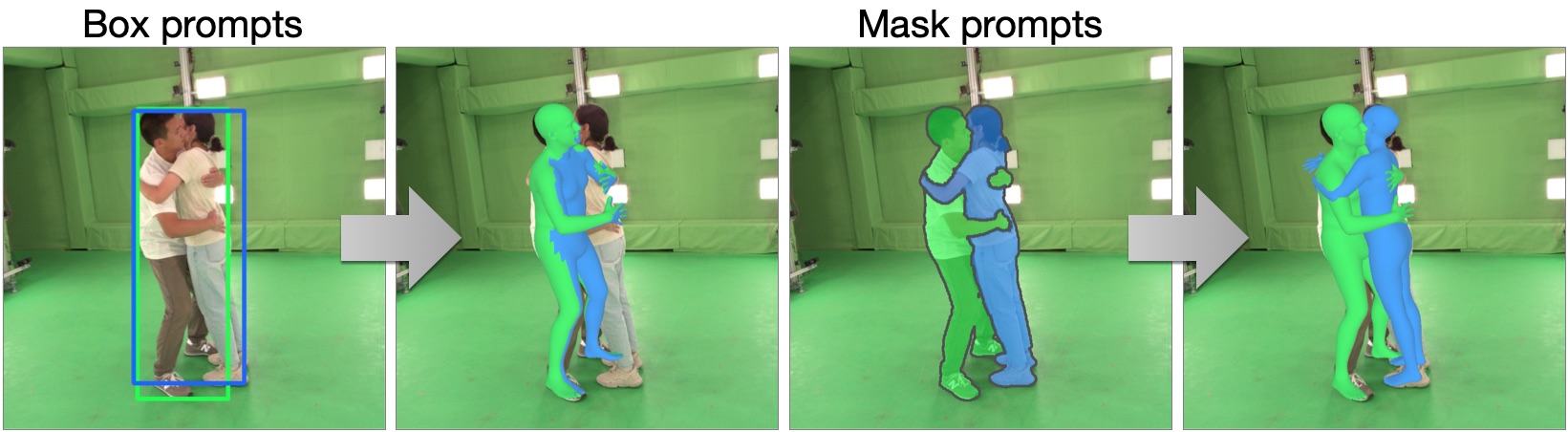}
    \vspace{-2mm}
    \caption{\textbf{Effect of mask prompts.} Results are from the same model with different prompt inputs. 
    Masks are better for close interaction scenarios where boxes are ambiguous.}
    \label{fig:masks}
    \vspace{-2mm}
\end{figure}

\begin{figure}[ht!]
    \centering
    \vspace{-2mm}
    \includegraphics[width=0.47\textwidth]{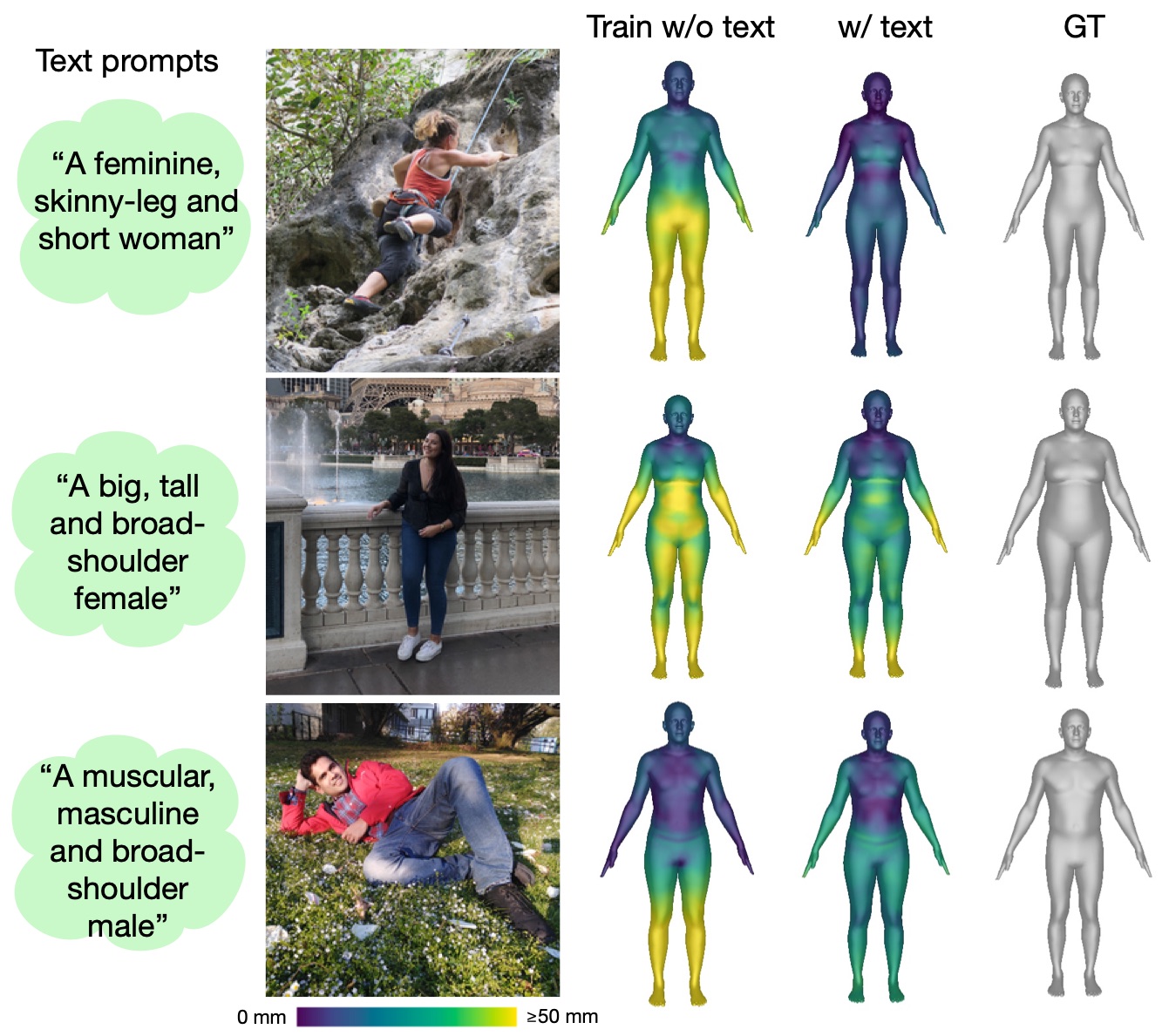}
    \vspace{-2mm}
    \caption{\textbf{Effect of shape prompts.} Compared to the baseline that does not incorporate shape description during training and testing, the model with shape prompts has better accuracy on HBW, especially in ambiguous images.}
    \label{fig:shape}
    \vspace{-2mm}
\end{figure}

\begin{table*}[ht!]
\centering
\setlength{\tabcolsep}{3pt}
\renewcommand{\arraystretch}{1.1}
\resizebox{0.85\textwidth}{!}
{\scriptsize{
\begin{tabular}{cl?ccc?ccc?ccc}
\cmidrule[0.75pt]{1-11}
& & \multicolumn{3}{c}{3DPW (14)} & \multicolumn{3}{c}{EMDB (24)} & \multicolumn{3}{c}{RICH (24)} \\
\cmidrule(lr){3-5} \cmidrule(lr){6-8} \cmidrule(lr){9-11}

& Models & \scriptsize{PA-MPJPE} & \scriptsize{MPJPE} & \scriptsize{PVE} & \scriptsize{PA-MPJPE} & \scriptsize{MPJPE} & \scriptsize{PVE} & \scriptsize{PA-MPJPE} & \scriptsize{MPJPE} & \scriptsize{PVE} \\
\cmidrule{1-11}

\multirow{4}{1em}{\rotatebox[origin=c]{90}{\tiny{cropped image}}} 
& CLIFF$^{\star}$~\cite{cliff} & 43.0 & 69.0 & 81.2 & 68.3 & 103.3 & 123.7 & 68.1 & 103.3 & 128.0  \\

& HMR2.0a~\cite{hmr2}  & 44.4 & 69.8 & 82.2 & 61.5 & 97.8 & 120.0 & 60.7 & 98.3 & 120.8 \\

& TokenHMR~\cite{tokenhmr}  & 44.3 & 71.0 &  84.6 & 55.6 & 91.7 & 109.4 & -- & -- & -- \\

& CameraHMR~\cite{camerahmr}  & 35.1 & 56.0 &  65.9 & 43.3 & 70.2 & 81.7 & 34.0 & 55.7 & 64.4 \\

\cmidrule{1-11}

\multirow{3}{1em}{\rotatebox[origin=c]{90}{\tiny{full image}}} 

& BEV~\cite{bev} & 46.9 & 78.5 & 92.3  & 70.9 & 112.2 & 133.4 & -- & -- & -- \\

& Multi-HMR$^{\star}$~\cite{multihmr}  & 45.9 & 73.1 & 87.1 & 50.1 & 81.6 & 95.7 & 46.3 & 73.8 & 83.0 \\

&\textbf{PromptHMR$^{\star}$} & \textbf{36.6} & \textbf{58.7} & \textbf{69.4} & \textbf{41.0} & \textbf{71.7} & \textbf{84.5} & \textbf{37.3} & \textbf{56.6} & \textbf{65.5} \\
\cmidrule{1-11}

\multirow{4}{1em}{\rotatebox[origin=c]{90}{\tiny{video}}} 

& WHAM~\cite{wham} & 37.5 & 59.8 & 71.5 & 52.0 & 81.6 & 96.9 & 44.3 & 80.0 & 91.2 \\
& TRAM~\cite{wang2024tram} & 35.6 & 59.3 & 69.6 & 45.7 & 74.4 & 86.6 & - & - & - \\
& GVHMR~\cite{shen2024gvhmr} & 37.0 & \textbf{56.6} & 68.7 & 44.5 & 74.2 & 85.9 & 39.5 & 66.0 & 74.4 \\
& \textbf{PromptHMR-Vid} & \textbf{35.5} & 56.9 & \textbf{67.3} & \textbf{40.1} & \textbf{68.1} & \textbf{79.2} & \textbf{37.0} & \textbf{57.4} & \textbf{65.8} \\
\cmidrule[0.75pt]{1-11}
\end{tabular}
}}
\vspace{-2mm}
\caption{\textbf{Comparison of mesh reconstruction} on the 3DPW, EMDB and RICH datasets, with the number of joints in parenthesis. $\star$ denotes methods that use ground truth focal length during inference. Note that we remove the test-time flip augmentation from all of the video methods to ensure a fair comparison. All metrics are in $mm$. }
\vspace{-3mm}
\label{tab:main_image}
\end{table*}

\begin{table}[ht!]
\centering
\setlength{\tabcolsep}{3pt}
\renewcommand{\arraystretch}{1.1}
\resizebox{0.47\textwidth}{!}
{\small{
\begin{tabular}{l?ccc?ccc}
\cmidrule[0.75pt]{1-7}
 & \multicolumn{3}{c}{HI4D (14)} & \multicolumn{3}{c}{CHI3D (14)} \\
\cmidrule(lr){2-4} \cmidrule(lr){5-7}

Models  & \scriptsize{PA-MPJPE} & \scriptsize{MPJPE} & \scriptsize{Pair-PA-MPJPE} & \scriptsize{PA-MPJPE} & \scriptsize{MPJPE} & \scriptsize{Pair-PA-MPJPE} \\
\cmidrule{1-7}

BEV$^{*}$~\cite{bev} & 81 & -- & 136 & 51 & -- & 96 \\
BUDDI~\cite{buddi} & 73 & -- & 98 & 47 &  -- & 68 \\
Multi-HMR$^{*}$~\cite{multihmr} & 49.8 & 67.8 & 80.6 & 31.7 & 54.0 & 100.0 \\
PromptHMR$^{*}$ & 39.2 & 63.9 & 78.1 & 27.2 & 48.0 & 58.5 \\
\textbf{PromptHMR} & \textbf{30.1} & \textbf{39.6} & \textbf{39.5} & \textbf{24.7} & \textbf{46.5} & \textbf{45.3} \\

\cmidrule[0.75pt]{1-7}
\end{tabular}
}}
\vspace{-2mm}
\caption{\textbf{Comparison on interaction reconstruction}. PromptHMR is more accurate in per-person and inter-person accuracy. $*$ denote a method or baseline is not trained on HI4D or CHI3D. All metrics are in $mm$. 
The impact of HI4D and the interaction prompt are evaluated in Table~\ref{tab:ablation_interaction}.
%Ablation of the contribution of the data and interaction prompts is included in Table~\ref{tab:ablation_interaction}.
}
\label{tab:main_interaction}
\vspace{-2mm}
\end{table}

\begin{table}[ht!]
\centering
\setlength{\tabcolsep}{3pt}
\renewcommand{\arraystretch}{1.1}
\resizebox{0.47\textwidth}{!}
{\scriptsize{
\begin{tabular}{cc?ccccc}
\cmidrule[0.75pt]{1-7}
\multirow{2}{*}{\vspace{-2mm}\makecell{Train \\ w/ text}} & \multirow{2}{*}{\vspace{-2mm}\makecell{Test \\ w/ text}} & \multicolumn{5}{c}{HBW} \\
\cmidrule(lr){3-7} 
& & Height & Chest & Waist & Hip & P2P-20k \\
\cmidrule{1-7}

$\times$ &  $\times$  & 69 & 51 & 88 & 63 & 26\\
$\surd$ &   $\times$      & 69 & 48 & 86 & 60 & 26 \\
$\surd$ & $\surd$ & \textbf{62} & \textbf{43} & \textbf{76} & \textbf{58} & \textbf{24} \\

\cmidrule[0.75pt]{1-7}
\end{tabular}
}}
\vspace{-2mm}
\caption{\textbf{Ablation of shape prompts using text}. Training with shape prompts improves shape accuracy. Using shape prompts during inference further improves shape accuracy. The ablation study is conducted with a 448$\times$448 model. Errors are in $mm$. }
\label{tab:ablation_shape}
\vspace{-1mm}
\end{table}

\begin{table}[ht!]
\centering
\setlength{\tabcolsep}{3pt}
\renewcommand{\arraystretch}{1.1}
\resizebox{0.47\textwidth}{!}
{\scriptsize{
\begin{tabular}{c?ccccc}
\cmidrule[0.75pt]{1-6}
\multicolumn{1}{c}{} & \multicolumn{5}{c}{EMDB-2 (24)} \\
\cmidrule(lr){2-6}
Models & $\text{WA-MPJPE}_{100}$ & $\text{W-MPJPE}_{100}$ & RTE & Jitter & Foot Skating \\
\cmidrule(lr){1-6} 
WHAM~\cite{wham} & 135.6 & 354.8 & 6.0 & 22.5 & 4.4 \\
TRAM~\cite{wang2024tram} & 76.4 & 222.4 & 1.4 & 18.5 & 23.4 \\
GVHMR~\cite{shen2024gvhmr} & 111.0 & 276.5 & 2.0 & 16.7 & \textbf{3.5} \\
\textbf{PromptHMR-Vid} & \textbf{71.0} & \textbf{216.5} & \textbf{1.3} & \textbf{16.3} & \textbf{3.5} \\
% \multirow{2}{*}{\vspace{-2mm}\makecell{Train \\ w/ text}} & \multirow{2}{*}{\vspace{-2mm}\makecell{Test \\ w/ text}} & \multicolumn{5}{c}{HBW} \\
% \cmidrule(lr){3-7} 
% & & Height & Chest & Waist & Hip & P2P-20k \\
% \cmidrule{1-7}

        % &         & 69 & 51 & 88 & 63 & 26\\
% $\surd$ &         & 70 & 48 & 83 & 61 & 26 \\
% $\surd$ & $\surd$ & 64 & 43 & 80 & 59 & 24 \\

\cmidrule[0.75pt]{1-6}
\end{tabular}
}}
\vspace{-2mm}
\caption{\textbf{Evaluation of motion in world coordinates}. PromptHMR-Vid combined with metric SLAM from TRAM~\cite{wang2024tram} surpasses SOTA methods at predicting human motion in world coordinates.}
\label{tab:world_coord_eval}
\vspace{-1mm}
\end{table}

\begin{table}[ht!]
\centering
\setlength{\tabcolsep}{3pt}
\renewcommand{\arraystretch}{1.1}
\resizebox{0.48\textwidth}{!}
{\small{
\begin{tabular}{ccc?ccc}
\cmidrule[0.75pt]{1-6}
\multicolumn{3}{c}{Trained with} & \multicolumn{3}{c}{HI4D (14)} \\
\cmidrule(lr){1-3}  \cmidrule(lr){4-6} 
Mask & Interaction & HI4D & PA-MPJPE & MPJPE & Pair-PA-MPJPE \\
\cmidrule{1-6}

$\times$ & $\times$ &  $\times$ & 47.0 & 71.4 & 87.2  \\
  $\surd$ & $\times$ &$\times$ & 43.4 & 60.5 & 83.0  \\
$\times$ & $\surd$ & $\times$ & 43.7 & 61.3 & 73.0  \\
$\times$  & $\times$ & $\surd$ & \textbf{36.3} & \underline{49.4} & \underline{52.6} \\
$\surd$ & $\surd$ & $\surd$ & \underline{36.5} & \textbf{47.1} & \textbf{47.9}  \\

\cmidrule[0.75pt]{1-6}
\end{tabular}
}}
\vspace{-2mm}
\caption{\textbf{Ablation on interaction prompt}. The interaction module improves inter-person reconstruction metrics Pair-PA-MPJPE on HI4D, especially when the method does not include HI4D in training. Ablation is conducted with a 448$\times$448 model. All metrics are in $mm$.}
\label{tab:ablation_interaction}
\vspace{-2mm}
\end{table}

% \begin{table}[ht!]
% \centering
% \setlength{\tabcolsep}{3pt}
% \renewcommand{\arraystretch}{1.1}
% \resizebox{0.48\textwidth}{!}
% {\small{
% \begin{tabular}{ccc?ccc}
% \cmidrule[0.75pt]{1-6}
% \multicolumn{3}{c}{Trained with} & \multicolumn{3}{c}{HI4D (14)} \\
% \cmidrule(lr){1-3}  \cmidrule(lr){4-6} 
% Interaction & CHI3D & HI4D & PA-MPJPE & MPJPE & Pair-PA-MPJPE \\
% \cmidrule{1-6}

%  & $\surd$ & & 47.0 & 71.4 & 87.2  \\
% $\surd$ & $\surd$ & & 43.7 & 61.3 & 73.0  \\
%  & $\surd$ & $\surd$ & 36.3 & 49.4 & 52.6 \\
% $\surd$ & $\surd$ & $\surd$ & 36.2 & 47.9 & 49.3  \\

% \cmidrule[0.75pt]{1-6}
% \end{tabular}
% }}
% \vspace{-2mm}
% \caption{\textbf{Ablation on interaction prompt}. The interaction module improves inter-person reconstruction metrics Pair-PA-MPJPE on HI4D, especially when the method does not include HI4D in training. Ablation is conducted with a 448$\times$448 model. All metrics are in $mm$.}
% \label{tab:ablation_interaction}
% \vspace{-2mm}
% \end{table}

\subsection{Effect of multimodal prompts}
We conduct qualitative and quantitative evaluations of the multimodal prompts. For efficient ablation, we train models with 448$\times$448 input resolution and select the best model within 150K steps of training.  

\begin{figure*}[h!]
    \centering
    \includegraphics[width=0.99\textwidth]{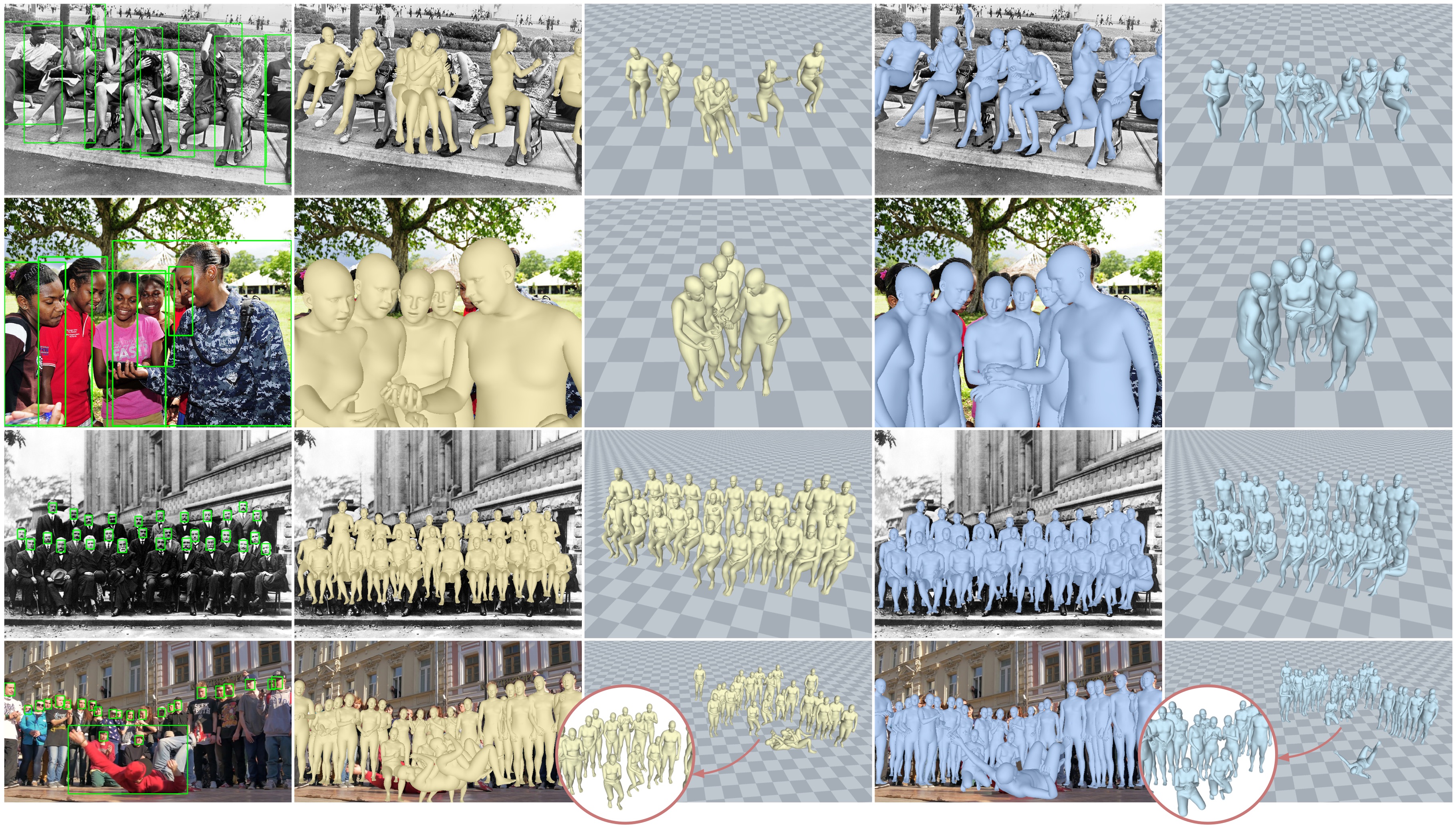}
    \vspace{-3mm}
    \caption{\textbf{Qualitative comparison:} Multi-HMR \raisebox{1mm}{\fcolorbox{black}{yellow!80}{\rule{1pt}{0pt}\rule{0pt}{1pt}}} vs PromptHMR \raisebox{1.mm}{\fcolorbox{black}{blue!60}{\rule{1pt}{0pt}\rule{0pt}{1pt}}}. Our model can recover coherent 3D scenes of people. In crowded scenes,  face detection provides reliable box prompts for our model. Please zoom in to see the details.}
    \label{fig:results}
\end{figure*}

For box prompts, as shown in rows 3-4 of Fig.~\ref{fig:results}, our method is able to take a combination of different boxes from in-the-wild images to reconstruct crowded scenes. Figure~\ref{fig:boxes} also shows an example with varying box inputs. PromptHMR remains stable when the boxes change and uses full image context to reconstruct the human even when the boxes are truncated.

The mask prompt is more effective than boxes when people closely overlap (Fig.~\ref{fig:masks}), as boxes are ambiguous in such cases. Ablation of HI4D (rows 1-2 in Tab.~\ref{tab:ablation_interaction}) shows that using  masks as the spatial prompt improves accuracy. 

Experiments on the HBW validation set (Tab.~\ref{tab:ablation_shape}) show that text prompts effectively improve shape accuracy when used during both training and testing. Moreover, training with shape descriptions alone provides an accuracy boost even if prompts are not given at test time. As illustrated in Fig.~\ref{fig:shape}, text prompts provide notable improvements, especially when large perspective effects create ambiguity.

For interaction prompts, we show an ablation in Table~\ref{tab:ablation_interaction}. The proposed interaction module is beneficial and largely improves inter-person accuracy on HI4D even without HI4D training, indicating out-off-domain generalization. When trained on HI4D, the interaction module does not improve per-person PA-MPJPE but still improves inter-person Pair-PA-MPJPE. Please refer to our SupMat for more qualitative results on interaction prompts.

\section{Limitations}
\label{sec:limits}
\vspace{-2mm}
We see PromptHMR as a step towards a holistic perception model for 3D humans, but several limitations need to be addressed in future work. Currently, the shape description and interaction prompts are not automatically generated and need to be supplied by the user. Future work should explore how to effectively integrate our promptable model with VLMs to automate prompting. We show how semantic prompts can improve reconstruction accuracy, but many other potential types of side information such as action descriptions, 3D scene context, or body measurements may provide additional benefits in different scenarios. 
\section{Conclusion}
\label{sec:conclusion}
We have presented PromptHMR, a promptable HPS estimation approach that leverages full image context with spatial and semantic prompts to infer 3D humans in the scene. Our method demonstrates state-of-the-art accuracy across diverse benchmarks and generalizes well in the wild. Our experiments show that incorporating diverse input information through flexible prompting enables robustness and adaptability in challenging scenarios.

\newpage
\textbf{Acknowledgement.} The authors would like to thank Yan Zhang, Yao Feng, and Nitin Saini for their suggestions. The majority of the work was done when Yufu was an intern at Meshcapade. Yufu and Kostas thank the support of NSF NCS-FO 2124355, NSF FRR 2220868, and NSF IIS-RI 2212433. 

\textbf{Disclosure.} While MJB is a co-founder and Chief Scientist at Meshcapade, his research in this project was performed solely at, and funded solely by, the Max Planck Society.

{
    \small
    \bibliographystyle{ieeenat_fullname}
    \bibliography{main}
}

\clearpage
\setcounter{page}{1}
\maketitlesupplementary

\section{Additional Results}
\label{sec:add_results}

In this section, we demonstrate more qualitative results to show the effects of interaction prompting and the video module. Please refer to the supplementary video to see the results from PromptHMR-Vid.

\subsection{Interaction Prompting}
\begin{figure*}[t!]
    \centering
    \vspace{-2mm}
    \includegraphics[width=0.98\textwidth]{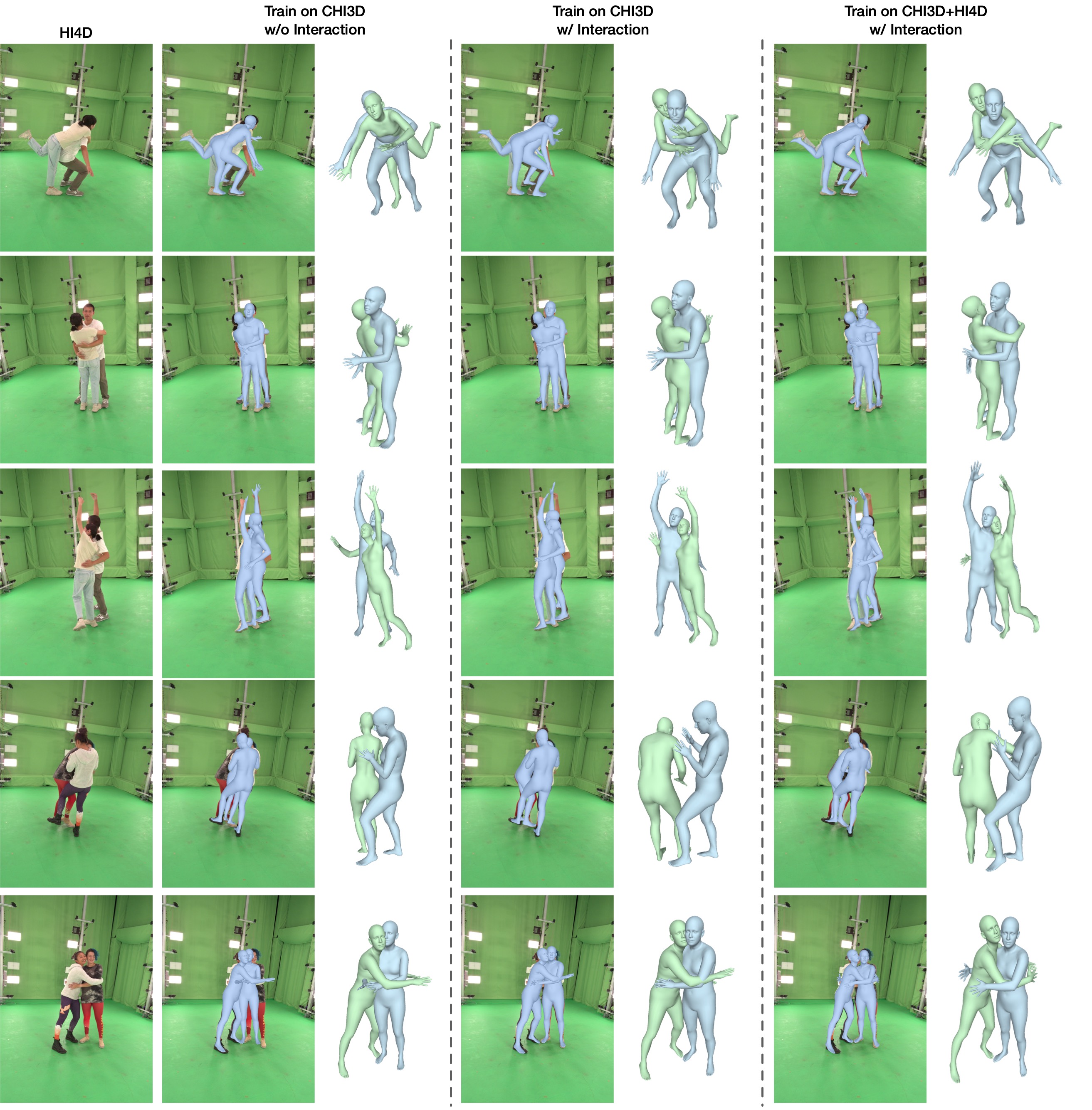}
    \vspace{-2mm}
    \caption{\textbf{Ablation of interaction module.} When fine-tuning the image model on CHI3D, adding the interaction module improves two-person interaction reconstruction on HI4D, which demonstrates the out-of-domain generalization ability of interaction prompting. Fine-tuning on both CHI3D and HI4D further improves results.}
    \label{fig:interaction_ablation}
    \vspace{-2mm}
\end{figure*}

We perform qualitative and quantitative ablation studies of interaction prompting on the HI4D dataset. 
In Tab. 5 of the main paper, we demonstrate that introducing interaction prompting improves the quantitative results on HI4D. 
In Fig.~\ref{fig:interaction_ablation}, we present more qualitative results to show the effect of the interaction module. As shown in the first column of Fig.~\ref{fig:interaction_ablation}, without the interaction module, the model does not learn to reconstruct close interaction effectively, even when trained with CHI3D interaction data. By adding the proposed interaction module, in the second column, the relative position and orientation of the interacting people are improved, and the penetration is reduced. Note that if we turn off the interaction module via the proposed flow control, the results will become similar to the first column. Finally, training with both CHI3D and HI4D leads to better results. 

\section{Experiment Details}
\label{sec:ex_details}
\subsection{Datasets}
The training set of the image model includes BEDLAM~\cite{bedlam}, AIC~\cite{aic}, InstaVariety~\cite{hmmr}, HI4D~\cite{hi4d}, CHI3D~\cite{chi3d}, AGORA~\cite{agora}, 3DPW~\cite{3dpw}, COCO~\cite{coco}, and MPII~\cite{mpii3d}, with the sampling rate of $\{0.2, 0.2, 0.3, 0.08, 0.08, 0.06, 0.06, 0.01, 0.01\}$. 
All input images are padded and resized to 896$\times$896. 
During training, we employ rotation and color jitter augmentation. For PromptHMR-Vid, we use BEDLAM and 3DPW datasets following~\cite{wham,shen2024gvhmr}.

To use datasets with different annotations for training, we adopt different losses described in Sec.3.5 of the main paper. 
For the ones (e.g. BEDLAM, AGORA, CHI3D, HI4D) with ground truth SMPL/SMPL-X annotations, we employ all loss items. 
While on AIC, InstaVariety, and 3DPW, we drop the translation loss. 
On COCO and MPII, we only compute 2D keypoint reprojection loss.

We generate the whole-body bounding boxes by projecting the ground-truth SMPL-X meshes onto the image plane. To generate the face bounding boxes, we project the head vertices. To generate truncated boxes, we take groups of keypoints (e.g. upper body keypoints) and compute their bounding boxes. Gaussian noise is then added to both corners.

On BEDLAM, AGORA, and AIC, we follow SHAPY~\cite{shapy} to compute the shape attribute scores. During training, we composite a shape description for each instance, such as ``a tall and broad-shoulder female" with a few augmentation rules. Each sentence will randomly sample 1-3 top attributes. The gender information is augmented with synonyms, such as ``female", ``woman", ``girl", etc.

\subsection{Architecture} 
We adopt the ViT-L~\cite{vit}, pretrained by DINOv2~\cite{oquab2023dinov2}, as our image encoder. We use an input image size of 896 and a patch size of 14, leading to the same spatial resolution as the recent Sapiens models~\cite{sapiens}. The text encoder is from MetaCLIP~\cite{metaclip}. The SMPL-X decoder consists of 3 attention blocks with an embedding dimension of 1024. From the output tokens ($T_{smpl}'$ and $T_{depth}'$), we use separate 2-layer MLPs to regress $\theta$, $\beta$, $p_{xy}$ and $p_{z}$ as introduced in Sec.3.2.

\subsection{Training}  
We train the PromptHMR image model using 8 H100 GPUs, with a batch size of 96 (12 images on each GPU). We use AdamW with a learning rate of 1e-5 for the image encoder, a learning rate of 3e-5 for the prompt encoder, and the SMPL-X decoder, $\beta_1$ of 0.9, $\beta_2$ of 0.999. and a weight decay of 5e-5. 

The losses presented in Sec.3.5 are weighted differently. For $\mathcal{L}_{2D}$, $\mathcal{L}_{3D}$, $\mathcal{L}_{\mathrm{SMPL}}$, $\mathcal{L}_{V}$ and $\mathcal{L}_{\mathit{transl}}$, the weights are set to $\{50.0, 5.0, 1.0, 1.0, 10.0\}$ respectively.

\noindent \paragraph{PromptHMR-Vid} We train the PromptHMR video model on 2 H100 GPUs with a batch size of 512 samples consisting of 120 frames each. We use AdamW with a learning rate of 2e-4 and a weight decay of 5e-5. We use the same losses as the image-based version in addition to binary cross-entropy loss for joint contact predictions.

\subsection{Metric}
In this section, we provide more details on the evaluation metric used in Sec.4 of the main paper. 

Mean Per Joint Position Error (\textbf{MPJPE}) is calculated by aligning the 3D joints obtained from SMPL-X with the ground truth at the pelvis before computing the mean square error. For historical reasons, different datasets use a different set of joints. Additionally, the pelvis definition could be different. To evaluate methods that predict SMPL-X on the datasets with SMPL labels, it's customary to convert the SMPL-X vertices to SMPL vertices and use a joint regressor on the converted vertices to obtain the 3D joints comparable to the labels. Note that all the above choices could alter the results and sometimes produce large ``artificial" improvements. So we strictly follow the most recent methods in the evaluation procedure. It's reported in the unit of $mm$.

Per Vertex error (\textbf{PVE}) computes mean square error on the vertices after pelvis alignment. Compared to MPJPE, it measures the combined pose and shape error. It's reported in the unit of $mm$.

Procrustes-aligned MPJPE (\textbf{PA-MPJPE}) performs general Procrustes alignment on the 3D joints before computing MPJPE. It measures purely the local articulated pose error. It's reported in the unit of $mm$.

Paired PA-MPJPE (\textbf{Pair-PA-MPJPE}) aligns two people as a whole with the ground truth before computing MPJPE. In addition to per-person error, it also measures the error in the relative position and orientation of the two people. It's used in HI4D and CHI3D to evaluate interaction reconstruction. It's reported in the unit of $mm$.

World-aligned MPJPE$_{100}$ (\textbf{WA-MPJPE$_{100}$}) measures the world-grounded motion accuracy. It aligns a segment of 100 frames of predictions with the ground truth before computing MPJPE. It's reported in the unit of $mm$.

World MPJPE$_{100}$ (\textbf{W-MPJPE$_{100}$}) is similar to WA-MPJPE but only aligns the first two frames of the 100-frame segment. Therefore, it provides a better measurement of the drifting in the direction and scale of the trajectories. It's reported in the unit of $mm$.

Root Trajectory Error (\textbf{RTE}) measures the accuracy of the whole trajectory including the scale. It performs \textbf{rigid alignment} on the trajectory of the root and computes the mean square error. It's reported in the unit of $\%$.

Motion Jitter (\textbf{Jitter}) uses finite difference to compute the jerk (3$^{rd}$ derivative) on the 3D joints. It measures rapid abrupt changes. It's reported in the unit of 10 $m/s^3$.

\textbf{Foot Skating} measures erroneous foot sliding. It thresholds the velocity of the ground truth foot vertices to compute contact frames, and calculates the displacement on the predicted foot vertices during contact. It's reported in the unit of $mm$.

\end{document}